\documentclass{article} 
\usepackage{nips12submit_e,times,amsmath,multirow,graphicx,cite}

\title{Feature Learning in Deep Neural Networks -- Studies on Speech Recognition Tasks}

\author{
Dong Yu, Michael L. Seltzer, Jinyu Li$^1$, Jui-Ting Huang$^1$, Frank Seide$^2$ \\
Microsoft Research, Redmond, WA 98052 \\
$^1$Microsoft Corporation, Redmond, WA 98052 \\
$^2$Microsoft Research Asia, Beijing, P.R.C.\\
\texttt{\{dongyu,mseltzer,jinyli,jthuang,fseide\}@microsoft.com} \\
}

\newcommand{\eqnref}[1]{(\ref{#1})}
\newcommand{\figref}[1]{Figure \ref{#1}}
\newcommand{\secref}[1]{Section \ref{#1}}
\newcommand{\tabref}[1]{Table \ref{#1}}

\nipsfinalcopy 

\begin{document}

\maketitle

\begin{abstract}
Recent studies have shown that deep neural networks (DNNs) perform significantly better than shallow networks and Gaussian mixture models (GMMs) on large vocabulary speech recognition tasks. In this paper, we argue that the improved accuracy achieved by the DNNs is the result of their ability to extract discriminative internal representations that are robust to the many sources of variability in speech signals. We show that these representations become increasingly insensitive to small perturbations in the input with increasing network depth, which leads to better speech recognition performance with deeper networks. We also show that DNNs cannot extrapolate to test samples that are substantially different from the training examples. If the training data are sufficiently representative, however, internal features learned by the DNN are relatively stable with respect to speaker differences, bandwidth differences, and environment distortion. This enables DNN-based recognizers to perform as well or better than state-of-the-art systems based on GMMs or shallow networks without the need for explicit model adaptation or feature normalization.

\end{abstract}

\section{Introduction}
\label{sec:intro}

Automatic speech recognition (ASR) has been an active research area for more than five decades. However, the performance of ASR systems is still far from satisfactory and the gap between ASR and human speech recognition is still large on most tasks.
One of the primary reasons speech recognition is challenging is the high variability in speech signals. For example, speakers may have different accents, dialects, or pronunciations, and speak in different styles, at different rates, and in different emotional states. The presence of environmental noise, reverberation, different microphones and recording devices results in additional variability. To complicate matters, the sources of variability are often nonstationary and interact with the speech signal in a nonlinear way. As a result, it is virtually impossible to avoid some degree of mismatch between the training and testing conditions.

Conventional speech recognizers use a hidden Markov model (HMM) in which each acoustic state is modeled by a Gaussian mixture model (GMM). The model parameters can be discriminatively trained  using an objective function such as maximum mutual information (MMI) \cite{MMI} or minimum phone error rate (MPE) \cite{povey2002mpe}. Such systems are known to be susceptible to performance degradation when even mild mismatch between training and testing conditions is encountered. To combat this, a variety of techniques has been developed. For example, mismatch due to speaker differences can be reduced by Vocal Tract Length Normalization (VTLN)\cite{zhan97}, which nonlinearly warps the input feature vectors to better match the acoustic model, or Maximum Likelihood Linear Regression (MLLR)\cite{gales1998mll}, which adapt the GMM parameters to be more representative of the test data. Other techniques such as Vector Taylor Series (VTS) adaptation are designed to address the mismatch caused by environmental noise and channel distortion \cite{acero00}. While these methods have been successful to some degree, they add complexity and latency to the decoding process. Most require multiple iterations of decoding and some only perform well with ample adaptation data, making them unsuitable for systems that process short utterances, such as voice search.

Recently, an alternative acoustic model based on deep neural networks (DNNs) has been proposed. In this model, a collection of Gaussian mixture models is replaced by a single context-dependent deep neural network (CD-DNN). A number of research groups have obtained strong results on a variety of large scale speech tasks using this approach \cite{YuDengDahl2010,Dahl2012,Seide:SWB-DNN,SeideLiChenYu2011,YuSeideLiDeng2012,Jaitly2012,SainathASRU2011,DahlICASSP2011}. Because the temporal structure of the HMM is maintained, we refer to these models as CD-DNN-HMM acoustic models.

In this paper, we analyze the performance of DNNs for speech recognition and in particular, examine their ability to learn representations that are robust to variability in the acoustic signal. To do so, we interpret the DNN as a joint model combining a nonlinear feature transformation and a log-linear classifier. Using this view, we show that the many layers of nonlinear transforms in a DNN convert the raw features into a highly invariant and discriminative representation which can then be effectively classified using a log-linear model. These internal representations become increasingly insensitive to small perturbations in the input with increasing network depth. In addition, the classification accuracy improves with deeper networks, although the gain per layer diminishes. However, we also find that DNNs are unable to extrapolate to test samples that are substantially different from the training samples. A series of experiments demonstrates that if the training data are sufficiently representative, the DNN learns internal features that are relatively invariant to sources of variability common in speech recognition such as speaker differences and environmental distortions.  This enables DNN-based speech recognizers to perform as well or better than state-of-the-art GMM-based systems without the need for explicit model adaptation or feature normalization algorithms.

The rest of the paper is organized as follows. In \secref{sec:dnn} we briefly describe DNNs and illustrate the feature learning interpretation of DNNs. In \secref{sec:invariant} we show that DNNs can learn invariant and discriminative features and demonstrate empirically that higher layer features are less sensitive to perturbations of the input. In \secref{sec:seeing} we point out that the feature generalization ability is effective only when test samples are small perturbations of training samples. Otherwise, DNNs perform poorly as indicated in our mixed-bandwidth experiments. We apply this analysis to speaker adaptation in \secref{sec:adaptation} and find that deep networks learn speaker-invariant representations, and to the Aurora 4 noise robustness task in \secref{sec:noise} where we show that a DNN can achieve performance equivalent to the current state of the art without requiring explicit adaptation to the environment. We conclude the paper in \secref{sec:conclude}.

\section{Deep Neural Networks}
\label{sec:dnn}

A deep neural network (DNN) is conventional multi-layer perceptron (MLP) with many hidden layers (thus deep). If the input and output of the DNN are denoted as $x$ and $y$, respectively, a DNN can be interpreted as a directed graphical model that approximates the posterior probability \mbox{$p_{y|x} (y=s|x)$} of a class $s$ given an observation vector $x$, as a stack of $(L+1)$ layers of log-linear models. The first $L$ layers model the posterior probabilities of hidden binary vectors $h^\ell$ given input vectors $v^\ell$. If $h^\ell$ consists of $N^\ell$ hidden units, each denoted as $h_j^\ell$, the posterior probability can be expressed as
\begin{equation*}
p^\ell (h^\ell |v^{\ell} )\,\,=\,\,\prod_{j=1}^{N^\ell} \frac{e^{z_j^\ell (v^{\ell} )\cdot h_j^\ell }}{e^{z_j^\ell (v^{\ell} )\cdot 1}+e^{z_j^\ell (v^{\ell} ) \cdot 0} },\,\,\,\,\,0 \le \ell < L
\end{equation*}
where $z^\ell (v^{\ell} )= (W^\ell )^T v^{\ell}+a^\ell$,
and $W^\ell$ and $a^\ell$ represent the weight matrix and bias vector in the $\ell$-th layer, respectively.  Each observation is propagated forward through the network, starting with the lowest layer $(v^0 = x)$ . The output variables of each layer become the input variables of the next, i.e. $v^{\ell+1} = h^\ell$. In the final layer, the class posterior probabilities are computed as a multinomial distribution
\begin{equation}
p_{y|x}(y=s|x)\,\,\,=\,\,\, p^L (y=s|v^{L} ) \,\,\,=\,\,\, \frac{e^{z_s^L (v^{L} ) }} { \sum_{s'} e^{z_{s'}^L (v^{L} ) } } \,\,\,=\,\,\, \textrm{softmax}_s\left(v^{L}\right). \label{eq:softmax}
\end{equation}
Note that the equality between $p_{y|x}(y=s|x)$ and $p^L (y=s|v^{L} )$ is valid by making a mean-field approximation \cite{Saul96meanfield} at each hidden layer.


In the DNN, the estimation of the posterior probability $p_{y|x} (y=s|x)$ can also be considered a two-step deterministic process. In the first step, the observation vector $x$ is transformed to another feature vector $v^L$ through $L$ layers of non-linear transforms.
In the second step, the posterior probability $p_{y|x}(y=s|x)$ is estimated using the log-linear model \eqnref{eq:softmax} given the transformed feature vector $v^L$. If we consider the first $L$ layers fixed, learning the parameters in the softmax layer is equivalent to training a conditional maximum-entropy (MaxEnt) model on features $v^L$. In the conventional MaxEnt model, features are manually designed \cite{YuDengAcero2009}. 
In DNNs, however, the feature representations are jointly learned with the MaxEnt model from the data. This not only eliminates the tedious and potentially erroneous process of manual feature extraction but also has the potential to automatically extract invariant and discriminative features, which are difficult to construct manually.

In all the following discussions, we use DNNs in the framework of the CD-DNN-HMM \cite{YuDengDahl2010,Dahl2012,Seide:SWB-DNN,SeideLiChenYu2011,YuSeideLiDeng2012} and use speech recognition as our classification task. The detailed training procedure and decoding technique for CD-DNN-HMMs can be found in  \cite{YuDengDahl2010,Dahl2012,Seide:SWB-DNN}.

\section{Invariant and discriminative features}
\label{sec:invariant}

\subsection{Deeper is better}
\label{ssec:deeper}
Using DNNs instead of shallow MLPs is a key component to the success of CD-DNN-HMMs. \tabref{tab:layers_vs_wer}, which is extracted from \cite{Seide:SWB-DNN}, summarizes the word error rates (WER) on the Switchboard (SWB) \cite{Switchboard1} Hub5'00-SWB test set. Switchboard is a corpus of conversational telephone speech. The system was trained using the 309-hour training set with labels generated by Viterbi alignment from a maximum likelihood (ML) trained GMM-HMM system. The labels correspond to tied-parameter context-dependent acoustic states called senones.
Our baseline WER with the corresponding discriminatively trained traditional GMM-HMM system is 23.6\%, while the best CD-DNN-HMM achives 17.0\%---a 28\% relative error reduction (it is possible
to further improve the DNN to a one-third reduction by realignment \cite{Seide:SWB-DNN}).

We can observe that deeper networks outperform shallow ones. The WER decreases as the number of hidden layers increases, using a fixed layer size of 2048 hidden units.
In other words, deeper models have stronger discriminative ability than shallow models.
This is also reflected in the improvement of the training criterion (not shown).
More interestingly, if architectures with an equivalent number of parameters are compared, the deep models consistently outperform the shallow models when the deep model is sufficiently wide at each layer. This is reflected in the right column of the table, which shows the performance for shallow networks with the same number of parameters as the deep networks in the left column. Even if we further increase the size of an MLP with a single hidden layer to about 16000 hidden units we can only achieve a WER of 22.1\%, which is significantly worse than the 17.1\% WER that is obtained using a $7$$\times$$2$k DNN under the same conditions. Note that as the number of hidden layers further increases, only limited additional gains are obtained and performance saturates after 9 hidden layers. The 9x2k DNN performs equally well as a 5x3k DNN which has more parameters. In practice, a tradeoff needs to be made between the width of each layer, the additional reduction in WER and the increased cost of training and decoding as the number of hidden layers is increased.

\begin{table}[t]
\caption{Effect of CD-DNN-HMM network depth on WER (\%) on Hub5'00-SWB using the 309-hour Switchboard training set. DBN pretraining is applied.}
\label{tab:layers_vs_wer}
\begin{center}
\begin{tabular}{cc|cc}
$L \times N$ & WER & $1\times N $ & WER \\
\hline
$1 \times 2\textrm{k}$  	& 24.2  &--                         &-- \\
$2 \times 2\textrm{k}$	& 20.4	&--	                        &-- \\
$3 \times 2\textrm{k}$	& 18.4	&--	                        &-- \\
$4 \times 2\textrm{k}$	& 17.8	&--	                        &-- \\
$5 \times 2\textrm{k}$	& 17.2	&$1 \times 3772$	        &22.5 \\
$7 \times 2\textrm{k}$	& 17.1	&$1 \times 4634$	        &22.6 \\
$9 \times 2\textrm{k}$	& 17.0	&--	                        &-- \\
$5 \times 3\textrm{k}$	& 17.0	&--	                        &-- \\
--				&--	&$1 \times 16\textrm{k}$	&22.1 \\
\end{tabular}
\end{center}
\end{table}

\subsection{DNNs learn more invariant features}
\label{ssec:more_invariant}
We have noticed that the biggest benefit of using DNNs over shallow models is that DNNs learn more invariant and discriminative features. This is because many layers of simple nonlinear processing can generate a complicated nonlinear transform. To show that this nonlinear transform is robust to small variations in the input features, let's assume the output of layer $l-1$, or equivalently the input to the layer $l$ is changed from $v^\ell$ to $v^\ell + \delta^\ell$, where $\delta^\ell$ is a small change. This change will cause the output of layer $l$, or equivalently the input to the layer $\ell+1$ to change by
\begin{equation*}
\delta^{\ell+1} = \sigma(z^\ell(v^\ell + \delta^\ell)) - \sigma(z^\ell(v^\ell)) \approx \textrm{diag}\left( \sigma'(z^\ell(v^\ell)) \right) (w^\ell)^T \delta^\ell.
\end{equation*}
The norm of the change $\delta^{\ell+1}$ is
\begin{align}
\Vert \delta^{\ell+1} \Vert & \approx \Vert \textrm{diag}(\sigma' (z^\ell (v^\ell )))((w^\ell )^T \delta^\ell ) \Vert \nonumber \\
& \le \Vert \textrm{diag}(\sigma' (z^\ell (v^\ell ))) (w^\ell )^T \Vert \Vert \delta^\ell \Vert  \nonumber \\
& = \Vert \textrm{diag}(v^{\ell+1} \circ (1-v^{\ell+1} )) (w^\ell )^T \Vert \Vert \delta^\ell \Vert \label{eq:norm}
\end{align}
where $\circ$ refers to an element-wise product.

Note that the magnitude of the majority of the weights is typically very small if the size of the hidden layer is large. For example, in a $6{\times}2\textrm{k}$ DNN trained using 30 hours of SWB data, 98\% of the weights in all layers except the input layer have magnitudes less than 0.5.



\begin{figure}[t]
\begin{center}
\includegraphics[width=4in]{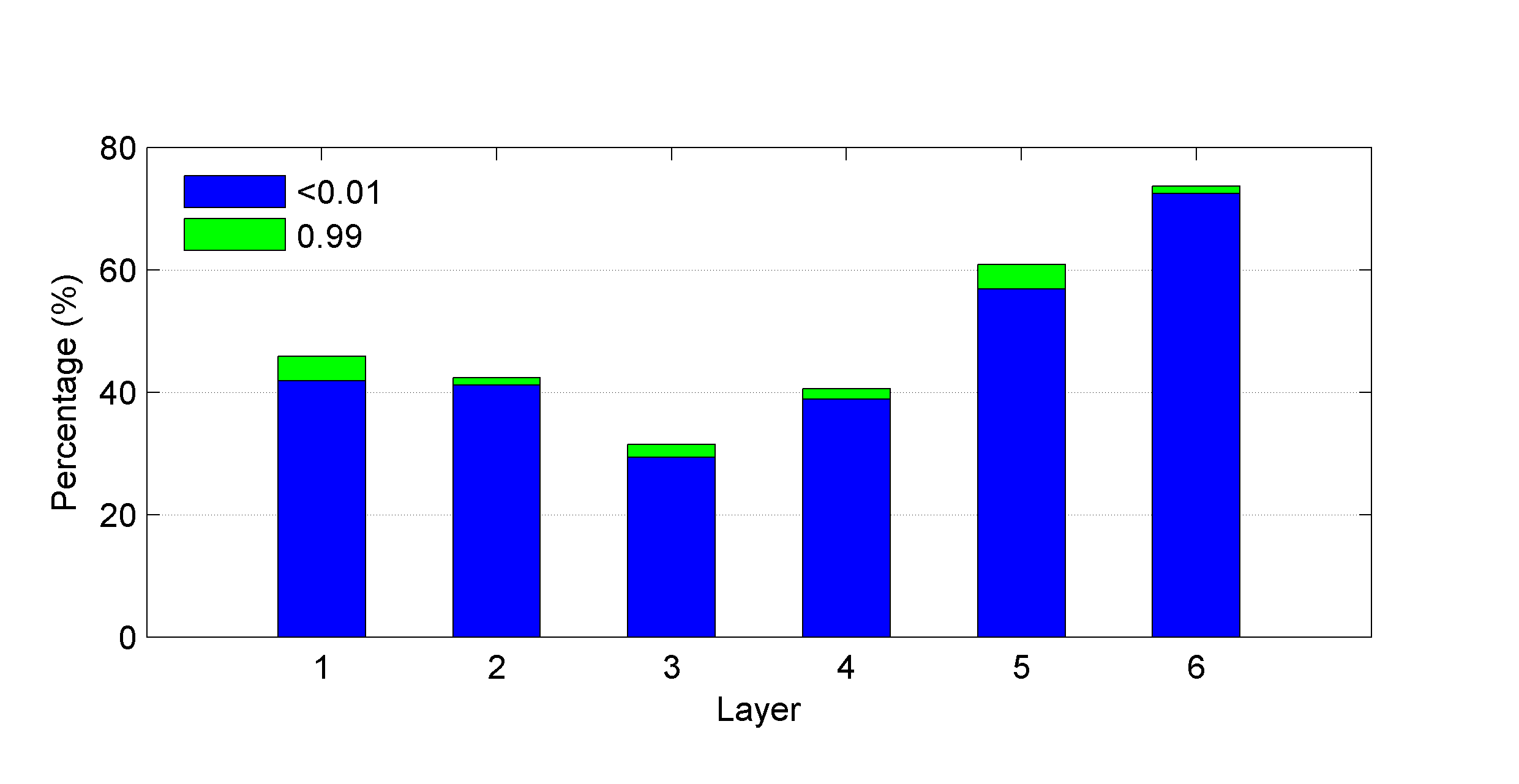}
\end{center}
\caption{Percentage of saturated activations at each layer}
\label{fig:saturation}
\end{figure}

\begin{figure}[t]
\begin{center}
\includegraphics[width=4in]{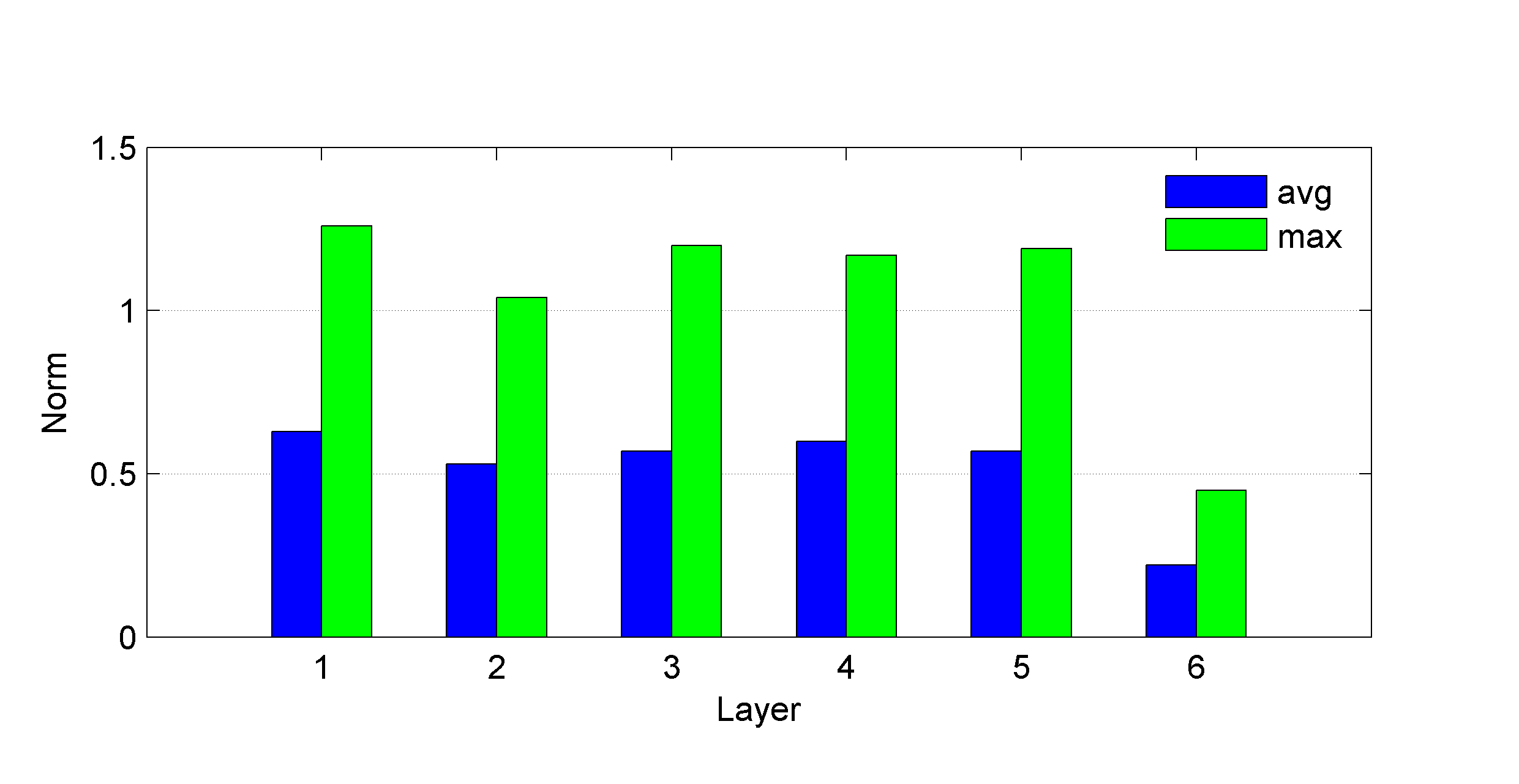}
\end{center}
\caption{Average and maximum $\Vert \textrm{diag}(v^{\ell+1} \circ (1-v^{\ell+1} )) (w^\ell )^T \Vert_2$ across layers on a $6{\times}2\textrm{k}$ DNN}
\label{fig:norm}
\end{figure}

While each element in $v^{\ell+1} \circ (1-v^{\ell+1} )$ is less than or equal to 0.25, the actual value is typically much smaller. This means that a large percentage of hidden neurons will not be active, as shown in \figref{fig:saturation}. As a result, the average norm $\Vert \textrm{diag}(v^{\ell+1} \circ (1-v^{\ell+1} )) (w^\ell )^T \Vert_2$ in \eqnref{eq:norm} across a 6-hr SWB development set is smaller than one in all layers, as indicated in \figref{fig:norm}. Since all hidden layer values are bounded in the same range of $(0, 1)$, this indicates that when there is a small perturbation on the input, the perturbation shrinks at each higher hidden layer. In other words, features generated by higher hidden layers are more invariant to variations than those represented by lower layers. Note that the maximum norm over the same development set is larger than one, as seen in \figref{fig:norm}. This is necessary since the differences need to be enlarged around the class boundaries to have discrimination ability.

\section{Learning by seeing}
\label{sec:seeing}

In \secref{sec:invariant}, we showed empirically that small perturbations in the input will be gradually shrunk as we move to the internal representation in the higher layers. In this section, we point out that the above result is only applicable to small perturbations around the training samples. When the test samples deviate significantly from the training samples, DNNs cannot accurately classify them. In other words, DNNs must see examples of representative variations in the data during training in order to generalize to similar variations in the test data.

We demonstrate this point using a mixed-bandwidth ASR study. Typical speech recognizers are trained on either narrowband speech signals, recorded at 8~kHz, or wideband speech signals, recorded at 16~kHz. It would be advantageous if a single system could recognize both narrowband and wideband speech, i.e. mixed-bandwidth ASR. One such system was recently proposed using a CD-DNN-HMM \cite{LiMixedBWDNN2012}. In that work, the following DNN architecture was used for all experiments. The input features were 29 mel-scale log filter-bank outputs together with dynamic features. An 11-frame context window was used generating an input layer with $29\cdot 3\cdot 11=957$ nodes. The DNN has 7 hidden layers, each with 2048 nodes. The output layer has 1803 nodes, corresponding to the number of senones determined by the GMM system.

\begin{figure}[t]
\begin{center}
\includegraphics[]{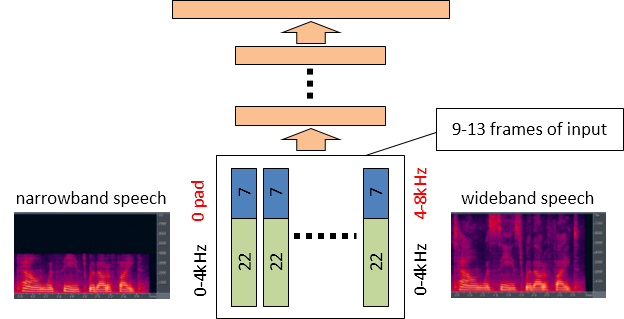}
\end{center}
\caption{Illustration of mixed-bandwidth speech recognition using a DNN}
\label{fig:mixedbwdnn}
\end{figure}

The 29-dimensional filter bank has two parts: the first 22 filters span 0--4~kHz and the last 7 filters span 4--8~kHz,  with the center frequency of the first filter in the higher filter bank at  4~kHz. When the speech is wideband, all 29 filters have observed values. However, when the speech is narrowband, the high-frequency information was not captured so the final 7 filters are set to 0. \figref{fig:mixedbwdnn} illustrates the architecture of the mixed-bandwidth ASR system.

Experiments were conducted on a mobile voice search (VS) corpus. This task consists of internet search queries made by voice on a smartphone.
There are two training sets, VS-1 and VS-2, consisting of 72 and 197 hours of wideband audio data, respectively. These sets were collected during different times of year. The test set, called VS-T, has 26757 words in 9562 utterances. The narrow band training and test data were obtained by downsampling the wideband data.

\tabref{tab:mixedbw_vs_wer} summarizes the WER on the wideband and narrowband test sets when the DNN is trained with and without narrowband speech. From this table, it is clear that if all training data are wideband, the DNN performs well on the wideband test set (27.5\% WER) but very poorly on the narrowband test set (53.5\% WER). However, if we convert VS-2 to narrowband speech and train the same DNN using mixed-bandwidth data (second row), the DNN performs very well on both wideband and narrowband speech.

\begin{table}[t]
\caption{WER (\%) on wideband (16k) and narrowband (8k) test sets with and without narrowband training data.}
\label{tab:mixedbw_vs_wer}
\begin{center}
\begin{tabular}{l|c|c}
training data 	& 16~kHz VS-T	& 8~kHz VS-T \\
\hline
16~kHz VS-1 + 16~kHz VS-2	& 27.5	& 53.5 \\
16~kHz VS-1 +  8~kHz VS-2  	& 28.3	& 29.3 \\
\end{tabular}
\end{center}
\end{table}

To understand the difference between these two scenarios, we take the output vectors at each layer for the wideband and narrowband input feature pairs,
$h^\ell(x_\textrm{wb})$ and $h^\ell(x_\textrm{nb})$,
and measure their Euclidean distance.
For the top layer, whose output is the senone posterior probability, we calculate the KL-divergence in nats
between $p_{y|x} (s_j |x_{\textrm{wb}} )$ and $p_{y|x} (s_j |x_{\textrm{nb}})$.
\tabref{tab:mixedbw_kl} shows the statistics of $d_l$ and $d_y$ over $40,000$ frames randomly sampled from the test set for the DNN trained using wideband speech only and the DNN trained using mixed-bandwidth speech.

\begin{table}[h]
\caption{Euclidean distance for the output vectors at each hidden layer (L1-L7) and the KL divergence (nats) for the posteriors at the top layer between the narrowband (8~kHz) and wideband (16~kHz) input features, measured using the wideband DNN or the mixed-bandwidth DNN.}
\label{tab:mixedbw_kl}
\begin{center}
\begin{tabular}{c|c|c|c|c|c}
 \multicolumn{2}{c|}{} & \multicolumn{2}{c|}{wideband DNN} & \multicolumn{2}{c}{mixed-band DNN} \\
 \hline
 layer & dist & mean & variance & mean & variance \\
\hline
L1	& \multirow{7}{*}{Eucl} & 13.28	& 3.90	& 7.32	& 3.62 \\
L2	&  & 10.38	& 2.47	& 5.39	& 1.28 \\
L3	&  & 8.04	& 1.77	& 4.49	& 1.27 \\
L4	&  & 8.53	& 2.33	& 4.74	& 1.85 \\
L5	&  & 9.01	& 2.96	& 5.39	& 2.30 \\
L6	&  & 8.46	& 2.60	& 4.75	& 1.57 \\
L7	&  & 5.27	& 1.85	& 3.12	& 0.93 \\
\hline
Top & KL & 2.03 & -- & 0.22 & -- \\
\end{tabular}
\end{center}
\end{table}

From \tabref{tab:mixedbw_kl} we can observe that in both DNNs, the distance between hidden layer vectors generated from the wideband and narrowband input feature pair is significantly reduced at the layers close to the output layer compared to that in the first hidden layer. Perhaps what is more interesting is that the average distances and variances in the data-mixed DNN are consistently smaller than those in the DNN trained on wideband speech only. This indicates that by using mixed-bandwidth training data, the DNN learns to consider the differences in the wideband and narrowband input features as irrelevant variations. These variations are suppressed after many layers of nonlinear transformation. The final representation is thus more invariant to this variation and yet still has the ability to distinguish between different class labels. This behavior is even more obvious at the output layer since the KL-divergence between the paired outputs is only 0.22 in the mixed-bandwidth DNN, much smaller than the 2.03 observed in the wideband DNN.

\section{Robustness to speaker variation}
\label{sec:adaptation}

A major source of variability is variation across speakers.
Techniques for adapting a GMM-HMM to a speaker have been investigated for decades.
Two important techniques are VTLN \cite{zhan97},
and feature-space MLLR (fMLLR) \cite{gales1998mll}.
%
%
%
Both VTLN and fMLLR operate on the features directly, making their application in the DNN context straightforward.

VTLN warps the frequency axis of the filterbank analysis to account for the fact that
the precise locations of vocal-tract resonances vary roughly monotonically with the physical size
of the speaker. This is done in both training and testing.
%
On the other hand, fMLLR applies an affine transform to the feature frames such that an adaptation data set better matches the model.
In most cases, including this work, `self-adaptation' is used: generate labels using unsupervised transcription, then re-recognize with the adapted model. This process is iterated four times. For GMM-HMMs, fMLLR transforms are estimated to maximize the likelihood of the adaptation data given the model.
For DNNs, we instead maximize cross entropy (with back propagation), which is a discriminative criterion,
so we prefer to call this transform feature-space Discriminative Linear Regression (fDLR). Note that the transform is applied to individual frames, prior to concatenation.
%

\def\cc#1{\multicolumn{1}{|c|}{#1}}
\def\ccl#1{\multicolumn{1}{|c}{#1}}

\begin{table}[t]
\caption{ {Comparison of feature-transform based speaker-adaptation techniques
for GMM-HMMs, a shallow, and a deep NN.
Word-error rates in \% for Hub5'00-SWB (relative change in parentheses).
}}
\label{tab:adaptation}
\begin{center}
\begin{tabular}{l|l|l|l}
\hline
					& \cc{GMM-HMM}	& \cc{CD-MLP-HMM}	& \ccl{CD-DNN-HMM}\\
adaptation technique			& \cc{40 mix}	& \cc{1$\times$2k}	& \ccl{7$\times$2k} \\
\hline
speaker independent			& 23.6		& 24.2		& 17.1		\\
\hline
+ VTLN					& 21.5 (-9\%)	& 22.5 (-7\%)	& 16.8 (-2\%)	\\
+ \{fMLLR/fDLR\}$\times$4		& 20.4 (-5\%)	& 21.5 (-4\%)	& 16.4 (-2\%)	\\
\hline
\end{tabular}
\end{center}
\end{table}

Typically, applying VTLN and fMLLR jointly to a GMM-HMM system will reduce errors by 10--15\%.
Initially, similar gains were expected for DNNs as well. However, these gains were not realized, as shown in \tabref{tab:adaptation} \cite{SeideLiChenYu2011}.
The table compares VTLN and fMLLR/fDLR for GMM-HMMs, a context-dependent ANN-HMM with a single hidden layer, and a deep network with 7 hidden layers,
on the same Switchboard task described in Section \ref{ssec:deeper}. For this task, test data are very consistent with the training, and thus, only a small amount of adaptation to other factors such as recording conditions or environmental factors occurs. We use the same configuration as in \tabref{tab:layers_vs_wer} which is speaker independent using single-pass decoding.

For the GMM-HMM, VTLN achieves a strong relative gain of 9\%. VTLN is also effective with the shallow neural-network system, gaining a slightly smaller 7\%. However, the improvement of VTLN on the deep network with 7 hidden layers is a much smaller 2\% gain.
%
Combining VTLN with fDLR further reduces WER by 5\% and 4\% relative, for the GMM-HMM and the shallow network, respectively. The reduction for the DNN is only 2\%.
%
We also tried transplanting VTLN and fMLLR transforms
estimated on the GMM system into the DNN, and achieved very similar results \cite{SeideLiChenYu2011}.

The VTLN and fDLR implementations of the shallow and deep networks are identical.
Thus, we conclude that to a significant degree, the deep neural network is able  to learn
internal representations that are invariant with respect to the sources of variability that VTLN and fDLR address.

\section{Robustness to environmental distortions}
\label{sec:noise}

In many speech recognition tasks, there are often cases where the despite the presence of variability in the training data, significant mismatch between training and test data persists. Environmental factors are common sources of such mismatch, e.g. ambient noise, reverberation, microphone type and capture device. 
The analysis in the previous sections suggests that DNNs have the ability to generate internal representations that are robust with respect to variability seen in the training data. In this section, we evaluate the extent to which this invariance can be obtained with respect to distortions caused by the environment.

We performed a series of experiments on the Aurora 4 corpus \cite{aurora4}, a 5000-word vocabulary task based on the Wall Street Journal (WSJ0) corpus. The experiments were performed with the 16~kHz multi-condition training set consisting of 7137 utterances from 83 speakers. One half of the utterances was recorded by a high-quality close-talking microphone and the other half was recorded using one of 18 different secondary microphones. Both halves include a combination of clean speech and speech corrupted by one of six different types of noise (street traffic, train station, car, babble, restaurant, airport) at a range of signal-to-noise ratios (SNR) between 10-20~dB.

The evaluation set consists of 330 utterances from 8 speakers. This test set was recorded by the primary microphone and a number of secondary microphones. These two sets are then each corrupted by the same six noises used in the training set at SNRs between 5-15~dB, creating a total of 14 test sets.  These 14 test sets can then be grouped into 4 subsets, based on the type of distortion: none (clean speech), additive noise only, channel distortion only, and noise + channel. Notice that the types of noise are common across training and test sets but the SNRs of the data are not.

The DNN was trained using 24-dimensional log mel filterbank features with utterance-level mean normalization. The first- and second-order derivative features were appended to the static feature vectors. The input layer was formed from a context window of 11 frames creating an input layer of 792 input units. The DNN had 7 hidden layers with 2048 hidden units in each layer and the final softmax output layer had 3206 units, corresponding to the senones of the baseline HMM system. The network was initialized using layer-by-layer generative pre-training and then discriminatively trained using back propagation.

In \tabref{tab:dnn_compare}, the performance obtained by the DNN acoustic model is compared to several other systems. The first system is a baseline GMM-HMM system, while the remaining systems are representative of the state of the art in acoustic modeling and noise and speaker adaptation. All used the same training set. To the authors' knowledge, these are the best published results on this task.

The second system combines Minimum Phone Error (MPE) discriminative training \cite{povey2002mpe} and noise adaptive training (NAT) \cite{Kalinli2010} using VTS adaptation to compensate for noise and channel mismatch \cite{flego2009discriminative}. The third system uses a hybrid generative/discriminative classifier \cite{ar527asru2011} as follows . First, an adaptively trained HMM with VTS adaptation is used to generate features based on state likelihoods and their derivatives. Then, these features are input to a discriminative log-linear model to obtain the final hypothesis. The fourth system uses an HMM trained with NAT and combines VTS adaptation for environment compensation and MLLR for speaker adaptation \cite{wang2012aslp}. Finally, the last row of the table shows the performance of the DNN system.

\begin{table}[h]
    \caption{A comparison of several systems in the literature to a DNN system on the Aurora 4 task.}
    \label{tab:dnn_compare}
    \begin{center}
    \begin{tabular}{l||c|c|c|c||c}
      \multirow{3}{*}{Systems}& \multicolumn{4}{c||}{distortion} & \multirow{3}{*}{AVG}\\
      \cline{2-5}
             & none  & \multirow{2}{*}{noise} & \multirow{2}{*}{channel} & noise + &   \\
             &(clean)&  &      & channel &   \\
        \hline
        GMM baseline                                       & 14.3 & 17.9 & 20.2 & 31.3 &  23.6 \\
        MPE-NAT + VTS\cite{flego2009discriminative}             & 7.2 & 12.8 & 11.5 & 19.7 & 15.3 \\
        NAT + Derivative Kernels \cite{ar527asru2011}     & 7.4 & 12.6 & 10.7 & 19.0 & 14.8 \\
        NAT + Joint MLLR/VTS  \cite{wang2012aslp}          & 5.6 & 11.0 &  8.8 & { 17.8} & 13.4 \\
        DNN ($7{\times}2048$)  & 5.6 &  8.8 &  8.9 & 20.0 & 13.4 \\
    \end{tabular}
    \end{center}
\end{table}

It is noteworthy that to obtain good performance, the GMM-based systems required complicated adaptive training procedures \cite{Kalinli2010,liao07} and multiple iterations of recognition in order to perform explicit environment and/or speaker adaptation. One of these systems required two classifiers. In contrast, the DNN system required only standard training and a single forward pass for classification. Yet, it outperforms the two systems that perform environment adaptation and matches the performance of a system that adapts to both the environment and speaker.

Finally, we recall the results in \secref{sec:seeing}, in which the DNN trained only on wideband data could not accurately classify narrowband speech. Similarly, a DNN trained only on clean speech has no ability to learn internal features that are robust to environmental noise. When the DNN for Aurora 4 is trained using only clean speech examples, the performance on the noise- and channel-distorted speech degrades substantially, resulting in an average WER of 30.6\%. This further confirms our earlier observation that DNNs are robust to modest variations between training and test data but perform poorly if the mismatch is more severe.

\section{Conclusion}
\label{sec:conclude}

In this paper we demonstrated through speech recognition experiments that DNNs can extract more invariant and discriminative features at the higher layers. In other words, the features learned by DNNs are less sensitive to small perturbations in the input features. This property enables DNNs to generalize better than shallow networks and enables CD-DNN-HMMs to perform speech recognition in a manner that is more robust to mismatches in speaker, environment, or bandwidth. On the other hand, DNNs cannot learn something from nothing. They require seeing representative samples to perform well. By using a multi-style training strategy and letting DNNs to generalize to similar patterns, we equaled the best result ever reported on the Aurora 4 noise robustness benchmark task without the need for multiple recognition passes and model adaptation.

\bibliographystyle{IEEEbib}
\renewcommand{\baselinestretch}{0.8}
\small
\bibliography{refs}

\begin{thebibliography}{10}

\bibitem{MMI}
L.~Bahl, P.~Brown, P.V. De~Souza, and R.~Mercer,
\newblock ``Maximum mutual information estimation of hidden markov model
  parameters for speech recognition,''
\newblock in {\em Proc. ICASSP}, Apr, vol.~11, pp. 49--52.

\bibitem{povey2002mpe}
D.~Povey and P.~C. Woodland,
\newblock ``Minimum phone error and i-smoothing for improved discriminative
  training,''
\newblock in {\em Proc. ICASSP}, 2002.

\bibitem{zhan97}
P.~Zhan et~al.,
\newblock ``Vocal tract length normalization for lvcsr,''
\newblock Tech. {R}ep. CMU-LTI-97-150, Carnegie Mellon Univ, 1997.

\bibitem{gales1998mll}
M.~J.~F. Gales,
\newblock ``{Maximum likelihood linear transformations for {HMM}-based speech
  recognition},''
\newblock {\em Computer Speech and Language}, vol. 12, pp. 75--98, 1998.

\bibitem{acero00}
A.~Acero, L.~Deng, T.~Kristjansson, and J.~Zhang,
\newblock ``{HMM Adaptation Using Vector Taylor Series for Noisy Speech
  Recognition},''
\newblock in {\em Proc. of ICSLP}, 2000.

\bibitem{YuDengDahl2010}
D.~Yu, L.~Deng, and G.~Dahl,
\newblock ``Roles of pretraining and fine-tuning in context-dependent
  {DBN-HMM}s for real-world speech recognition,''
\newblock in {\em Proc. NIPS Workshop on Deep Learning and Unsupervised Feature
  Learning}, 2010.

\bibitem{Dahl2012}
G.E. Dahl, D.~Yu, L.~Deng, and A.~Acero,
\newblock ``Context-dependent pretrained deep neural networks for large
  vocabulary speech recognition,''
\newblock {\em {IEEE} Trans. Audio, Speech, and Lang. Proc.}, vol. 20, no. 1,
  pp. 33--42, Jan. 2012.

\bibitem{Seide:SWB-DNN}
F.~Seide, G.~Li, and D.~Yu,
\newblock ``Conversational speech transcription using context-dependent deep
  neural networks,''
\newblock in {\em Proc. Interspeech}, 2011.

\bibitem{SeideLiChenYu2011}
F.~Seide, G.Li, X.~Chen, and D.~Yu,
\newblock ``Feature engineering in context-dependent deep neural networks for
  conversational speech transcription,''
\newblock in {\em Proc. ASRU}, 2011, pp. 24--29.

\bibitem{YuSeideLiDeng2012}
D.~Yu, F.~Seide, G.Li, and L.~Deng,
\newblock ``Exploiting sparseness in deep neural networks for large vocabulary
  speech recognition,''
\newblock in {\em Proc. ICASSP}, 2012, pp. 4409--4412.

\bibitem{Jaitly2012}
N.~Jaitly, P.~Nguyen, A.~Senior, and V.~Vanhoucke,
\newblock ``An application of pretrained deep neural networks to large
  vocabulary conversational speech recognition,''
\newblock Tech. {R}ep. Tech. Rep. 001, Department of Computer Science,
  University of Toronto, 2012.

\bibitem{SainathASRU2011}
T.~N. Sainath, B.~Kingsbury, B.~Ramabhadran, P.~Fousek, P.~Novak, and
  A.~r.~Mohamed,
\newblock ``Making deep belief networks effective for large vocabulary
  continuous speech recognition,''
\newblock in {\em Proc. ASRU}, 2011, pp. 30--35.

\bibitem{DahlICASSP2011}
G.~E. Dahl, D.~Yu, L.~Deng, and A.~Acero,
\newblock ``Large vocabulary continuous speech recognition with
  context-dependent dbn-hmms,''
\newblock in {\em Proc. ICASSP}, 2011, pp. 4688--4691.

\bibitem{Saul96meanfield}
L.~Saul, T.~Jaakkola, and M.~I. Jordan,
\newblock ``Mean field theory for sigmoid belief networks,''
\newblock {\em Journal of Artificial Intelligence Research}, vol. 4, pp.
  61--76, 1996.

\bibitem{YuDengAcero2009}
D.~Yu, L.~Deng, and A.~Acero,
\newblock ``Using continuous features in the maximum entropy model,''
\newblock {\em Pattern Recognition Letters}, vol. 30, no. 14, pp. 1295--1300,
  2009.

\bibitem{Switchboard1}
J.~Godfrey and E.~Holliman,
\newblock {\em Switchboard-1 Release 2},
\newblock Linguistic Data Consortium, Philadelphia, PA, 1997.

\bibitem{LiMixedBWDNN2012}
J.~Li, D.~Yu, J.-T. Huang, and Y.~Gong,
\newblock ``Improving wideband speech recognition using mixed-bandwidth
  training data in {CD-DNN-HMM},''
\newblock in {\em Proc. SLT}, 2012.

\bibitem{aurora4}
N.~Parihar and J.~Picone,
\newblock ``{Aurora working group: DSR front end LVCSR evaluation AU/384/02},''
\newblock Tech. {R}ep., Inst. for Signal and Information Process, Mississippi
  State University.

\bibitem{Kalinli2010}
O.~Kalinli, M.~L. Seltzer, J.~Droppo, and A.~Acero,
\newblock ``Noise adaptive training for robust automatic speech recognition,''
\newblock {\em IEEE Trans. on Audio, Sp. and Lang. Proc.}, vol. 18, no. 8, pp.
  1889 --1901, Nov. 2010.

\bibitem{flego2009discriminative}
F.~Flego and M.~J.~F. Gales,
\newblock ``{Discriminative adaptive training with VTS and JUD},''
\newblock in {\em Proc. ASRU}, 2009.

\bibitem{ar527asru2011}
A.~Ragni and M.~J.~F. Gales,
\newblock ``{Derivative kernels for noise robust ASR},''
\newblock in {\em Proc. ASRU}, 2011.

\bibitem{wang2012aslp}
Y.-Q. Wang and M.~J.~F. Gales,
\newblock ``Speaker and noise factorisation for robust speech recognition,''
\newblock {\em IEEE Trans. on Audio Speech and Language Proc.}, vol. 20, no. 7,
  2012.

\bibitem{liao07}
H.~Liao and M.~J.~F. Gales,
\newblock ``{Adaptive training with joint uncertainty decoding for robust
  recognition of noisy data},''
\newblock in {\em Proc. of ICASSP}, Honolulu, Hawaii, 2007.

\end{thebibliography}

\end{document}